# STELLAR: Siamese Multi-Headed Attention Neural Networks for Overcoming Temporal Variations and Device Heterogeneity with Indoor Localization

Danish Gufran, *Student Member, IEEE*, Saideep Tiku, *Member, IEEE*,
and Sudeep Pasricha, *Senior Member, IEEE*

*Abstract*— Smartphone-based indoor localization has emerged as a cost-effective and accurate solution to localize mobile and IoT devices indoors. However, the challenges of device heterogeneity and temporal variations have hindered its widespread adoption and accuracy. Towards jointly addressing these challenges comprehensively, we propose STELLAR, a novel framework implementing a contrastive learning approach that leverages a Siamese multi-headed attention neural network. STELLAR is the first solution that simultaneously tackles device heterogeneity and temporal variations in indoor localization, without the need for re-training the model (re-calibration-free). Our evaluations across diverse indoor environments show 8-75% improvements in accuracy compared to state-of-the-art techniques, to effectively address the device heterogeneity challenge. Moreover, STELLAR outperforms existing methods by 18-165% over 2 years of temporal variations, showcasing its robustness and adaptability.

## I. INTRODUCTION

IN recent years, indoor localization has garnered significant attention due to its wide range of applications, including healthcare, asset tracking, navigation systems, location-based advertising, and much more [1]. The ability to accurately determine the position of mobile and IoT devices within indoor environments has the potential to revolutionize various industries and enhance user experience [2]. However, achieving high-precision indoor localization remains a challenging task, due to the inherent complexities and dynamic nature of indoor environments.

To achieve high-precision indoor localization, many researchers in the domain have turned to Wi-Fi received signal strength (RSS) based indoor localization [3]. This method leverages using Wi-Fi RSS from Wi-Fi routers or access points (APs) to determine the position of mobile and IoT devices within indoor environments. Fingerprinting model-based systems [4] have gained significant popularity due to their robustness and compatibility with existing Wi-Fi infrastructure in complex indoor environments [5]. Despite their effectiveness, these systems have limitations. Inaccuracies may arise due to signal variability caused by obstacles, signal interference, and other factors, such as signal attenuation, multi-path fading, and shadowing, which can distort signals and result in inaccurate location estimates [6]. These limitations underscore the importance of developing robust localization methods to overcome these challenges.

Fingerprinting model-based systems typically involve a database of Wi-Fi signal patterns, commonly known as "fingerprints", are captured across the indoor space. This database allows for more robust and accurate positioning or localization [7], making it an increasingly preferred choice in recent indoor localization developments and competitions (e.g., [20], [21]), avoiding the need for additional hardware deployment [8]. However, changes in the indoor environment such as furniture and equipment rearrangement or the addition of new obstacles can render the stored fingerprints less accurate, necessitating frequent updates. Moreover, fingerprinting-based systems may face challenges in handling large-scale deployments, as the collection and management of extensive fingerprint databases can become cumbersome [7], [8]. Hence, there is a need for a robust approach for fingerprinting-based localization.

Machine learning (ML)-based fingerprinting approaches can be used to learn patterns and relationships from the collected RSS fingerprint data, enabling accurate positioning of devices in indoor environments [9]. ML-based fingerprinting systems consists of two phases: The offline (training) phase and the online (inferencing) phase [10]. During the offline or training phase, a dataset of RSS fingerprints is collected from known locations within the indoor environment. This dataset serves as the training data for the ML model. Once the ML model is trained, it can be used in the online or inference phase for real-time positioning

This work was supported in part by the National Science Foundation grant CNS-2132385. *(Corresponding author: D. Gufran).*
Danish Gufran is with the Department of Electrical and Computer Engineering, Colorado State University, Fort Collins, CO 80523 USA (e-mail: Danish.Gufran@colostate.edu).
Saideep Tiku is with Micron Semiconductor Products, Folsom, California. (e-mail: saideeptiku@micron.com).
Sudeep Pasricha is with the Department of Electrical and Computer Engineering, Colorado State University, Fort Collins, CO 80523 USA (e-mail: sudeep@colostate.edu).

of devices [5]-[13].

Unfortunately, ML-based fingerprinting systems face many challenges arising from the phenomenon of device heterogeneity. Mobile devices from different manufacturers use varying Wi-Fi chipset configurations, leading to different captured RSS patterns across devices that are at the same location. This discrepancy in captured RSS makes it difficult for the ML model to accurately predict the device's location [14]. Furthermore, indoor environments are not only subject to device heterogeneity but are also influenced by various sources of external noise [15]-[18]. Additionally, the often-overlooked temporal variations caused by changes in APs and indoor environments over months to years exacerbates the complexity of the localization process [15]-[18]. APs may be added, dropped, or turned off due to maintenance, power outages, or other reasons, leading to temporal variations that significantly impact the availability and quality of RSS signals [19]. Existing methods often struggle to adapt to these changes and maintain accuracy in the face of such challenges.

In this paper, we introduce STELLAR, an innovative framework that pioneers addressing both device heterogeneity and temporal resilience in indoor localization, all without requiring frequent model re-training ("re-calibration-free"). Unlike many ML-based fingerprinting methods, which typically require re-training to handle dynamic indoor challenges. STELLAR excels by not needing frequent recalibration or updates, ensuring consistent performance in the face of evolving indoor environments. By leveraging contrastive learning, an ML technique that maximizes similarity between similar RSS samples and minimizes it between dissimilar ones, STELLAR is able to address complex indoor localization challenges, resulting in a robust and long-term indoor localization solution. The novel contributions of this work are:

- We propose a framework called STELLAR to address both device heterogeneity and temporal variations, for accurate and long-term indoor localization.
- We develop a contrastive learning and re-calibration-free localization approach based on a Siamese multi-headed attention neural network ML model.
- Towards the goal of improving both device heterogeneity resilience and temporal invariance, we identify and adapt data modulation methodologies for training the proposed ML model.
- By collecting fingerprint(s) with multiple heterogeneous devices across buildings and time (few seconds to years), we create benchmarks to test the localization accuracy of STELLAR against the state-of-the-art.
- We prototype our re-calibration-free indoor localization framework, deploy it on diverse smartphones, and evaluate it under real-world settings.

In the following sections, we provide a detailed overview of the related work in indoor localization, discuss the methodology and architecture of the proposed STELLAR framework, present our experimental setup and results, and conclude with the implications of this research.

## II. RELATED WORKS

Indoor localization has been studied for several decades now, with Wi-Fi fingerprinting based indoor localization gaining prominence in recent competitions by Microsoft [20] and NIST [21]. Wi-Fi fingerprinting methods have shown better accuracy compared with other approaches, making fingerprinting a particularly promising approach for indoor positioning [8]. Simultaneously, advancements in the computational capabilities of smartphones have played a pivotal role in empowering them to execute deeper and more powerful ML models, which have the potential to enhance indoor localization performance. As a result, researchers have been actively exploring the integration of ML with Wi-Fi fingerprinting to improve indoor localization accuracy and efficiency in recent years.

Various ML algorithms, including support vector machines (SVM) [12], K-nearest neighbors (KNN) [22], random forest (RF) [23], and neural networks such as deep neural networks (DNN) [24] and convolutional neural networks (CNN) [25], have been employed to capture intricate non-linear relationships within indoor localization data. These ML algorithms, with their ability to handle complex patterns and relationships, have proven to deliver higher accuracies compared to traditional non-ML based fingerprinting-based solutions. However, despite the integration of ML techniques, fingerprinting-based methods may encounter challenges in handling device heterogeneity (fluctuations in RSS fingerprints due to various devices) and temporal variations (changes in APs over months to years indoors), as discussed earlier.

To address the device heterogeneity challenge, researchers have made use of advanced neural network-based ML algorithms in recent work. For instance, ANVIL [15] utilized a multi-headed attention neural network combined with domain-specific augmentation to address device heterogeneity in challenging indoor environments with signal reflections and obstructions. The use of attention mechanisms enhances the localization process in ANVIL by selectively focusing on relevant signal patterns. VITAL [26] introduces a method for converting Wi-Fi fingerprints into equivalent images, enabling the use of the vision transformer model with a unique image-based data augmentation module to tackle the challenge of device heterogeneity. The inclusion of an attention neural network within the vision transformer further improves localization accuracy by focusing on relevant signal patterns. VITAL demonstrates comparable performance to ANVIL [15] when addressing device heterogeneity for indoor localization. WiDeep [28] incorporates an autoencoder-supported Gaussian process classifier to tackle the challenges of device heterogeneity. By utilizing an autoencoder, WiDeep learns condensed and informative representations of Wi-Fi signal patterns, mitigating the impact of device variations and enhancing the model's generalization capabilities. Subsequently, the Gaussian process classifier harnesses these learned representations to effectively model the intricate non-linear relationships in the data. SANGRIA [29] employs a stacked autoencoder to address the challenges of device heterogeneity. SANGRIA outperforms WiDeep [28] due to its stacked autoencoder being able to learn hierarchical representations of Wi-Fi signal patterns more efficiently, enabling improved handling of device variations. While all of these works show promising potential, they do not address the challenges associated with temporal variations (changes in APs and indoor environments over months to years indoors), resulting in limited resilience over time.

A few recent works have begun to address temporal variations during indoor localization. LT-KNN (long-term KNN) [30] addresses situations where multiple APs are removed during the online phase (over a 15-month span of time), by incorporating regular re-training, thereby improving the performance of KNN in such scenarios. However, the need for frequent re-training due to sensitivity to temporal variations is a significant overhead. Another approach in [31] utilizes a simple Siamese neural network using fully connected layers to address the challenges associated with temporal variations. However, like LT-KNN [30], it suffers from the drawback of requiring frequent re-training. This re-training (or re-calibration) phase is necessary to adapt ML models to the dynamic changes in the indoor environment. But re-training requires collecting new data, which can be time-consuming, leading to temporary service disruptions and potential model overfitting or underfitting. These factors hinder the system's real-time responsiveness and adaptability, making it less practical for long-term indoor localization deployments.

Our proposed STELLAR framework described in this paper is a re-calibration-free framework that does not require re-training over time, while being able to simultaneously address both device heterogeneity and temporal variations. This framework builds upon our prior work ANVIL [15] that was presented at the 2022 IEEE IPIN conference. STELLAR improves upon ANVIL in several ways: (*i*) STELLAR addresses both device heterogeneity and temporal variations, whereas ANVIL only focuses on device heterogeneity, (*ii*) STELLAR utilizes a new gradient boosting approach to enhance resilience towards heterogeneity, leading to improved localization performance compared to ANVIL, (*iii*) unlike ANVIL, STELLAR is a re-calibration-free model, by leveraging a more sophisticated Siamese multi-headed attention approach for enhanced localization performance, and (*iv*) we add multiple new experiments and incorporate newer smartphone devices during the online phase to more comprehensively evaluate STELLAR's effectiveness for robust indoor localization.

### III. ANALYSIS OF DEVICE HETEROGENEITY VARIATIONS

Wi-Fi fingerprinting based indoor localization systems heavily rely on Wi-Fi RSS values captured by mobile devices for their accuracy and reliability. However, the presence of device heterogeneity introduces significant variations in these RSS readings, which can negatively impact localization results. To understand the implications of device heterogeneity on indoor localization accuracy, we conducted a comprehensive analysis using popular smartphones.

Wi-Fi RSS measurements reflect the power level of the wireless signal received from a Wi-Fi AP by a Wi-Fi receiver in devices such as smartphones, tablets, and laptops. These measurements are expressed in decibels relative to a milliwatt (dBm) and typically range from -100dBm to 0dBm where -100dBm indicates no signal or invisible AP and 0dBm represents the strongest signal [32]. Weak signals are usually defined as below -80dBm, while strong signals are considered to be above -40dBm. However, it is important to note that the precise threshold values for strong and weak RSS signals can vary based on factors such as Wi-Fi chipset, device firmware, and the surrounding environment [32].

In practice, capturing RSS values in real-world settings is not as straightforward as in an idealized scenario. Multiple factors, such as signal attenuation, reflections from objects, human interferences, and multipath propagation, can introduce variations in the observed RSS values. Moreover, the presence of device heterogeneity further complicates the situation. Variations in hardware, software, and antenna configurations among different mobile devices, even with the same underlying Wi-Fi chipset (from the same manufacturer), can lead to significant discrepancies in the reported RSS values [32]. Firmware configurations and signal processing algorithms also contribute to differences in RSS readings. As a result, training ML models can be challenging as heterogeneous and noisy RSS data can result in poor generalization and inaccurate localization predictions.

To investigate the influence of device heterogeneity, we captured Wi-Fi RSS values using three different smartphones namely, OnePlus 3, Samsung Galaxy S7, and LG V20, at the same indoor location and at the same time in a building. The observations are shown in Figure 1. Comparing the RSS values captured by each smartphone for various Wi-Fi APs, we can observe significant variations, that show the impact of device heterogeneity. For instance, when comparing the average RSS values for the AP with the first MAC ID '38:17:c3:1f:0e:10', Samsung Galaxy S7 recorded -47 dBm, OnePlus 3 recorded -41 dBm, and LG V20 recorded -79 dBm (on average). Similarly, for the AP with the MAC ID '80:8d:b7:55:35:11', Samsung Galaxy S7 captured -71 dBm, OnePlus 3 recorded -90 dBm, and LG V20 recorded -80 dBm (on average). Cumulative variations in hardware, firmware, and signal processing algorithms contribute to these differences and must be addressed in ML-based indoor localization systems to preserve accuracy across heterogeneous devices.

### IV. ANALYSIS OF TEMPORAL VARIATIONS

Temporal variations in indoor localization pose challenges due to changes in Wi-Fi APs or the indoor environment over time. Temporal variations can be attributed to several factors within the indoor environment, encompassing both short-term and long-term variations. Moreover, changes in AP configurations, such as APs being turned off, added, or dropped, lead to inconsistencies in RSS fingerprints and can impact the quality and availability of localization data, resulting in inaccurate predictions. Additionally, temporal variations can be categorized into short-term and long-term variations, each with its distinct impact on indoor localization.

Short-term variations encompass rapid and transient changes within the indoor space. For instance, the movement of people, conversations in specific areas, or even adjustments like turning screens or opening metal cabinet doors can significantly influence local RSS values, introducing fluctuations in the data.

Long-term variations, on the other hand, involve more persistent alterations in the indoor environment. These can include structural changes like modifications in the floorplan or furniture layout, as well as the installation or removal of APs.

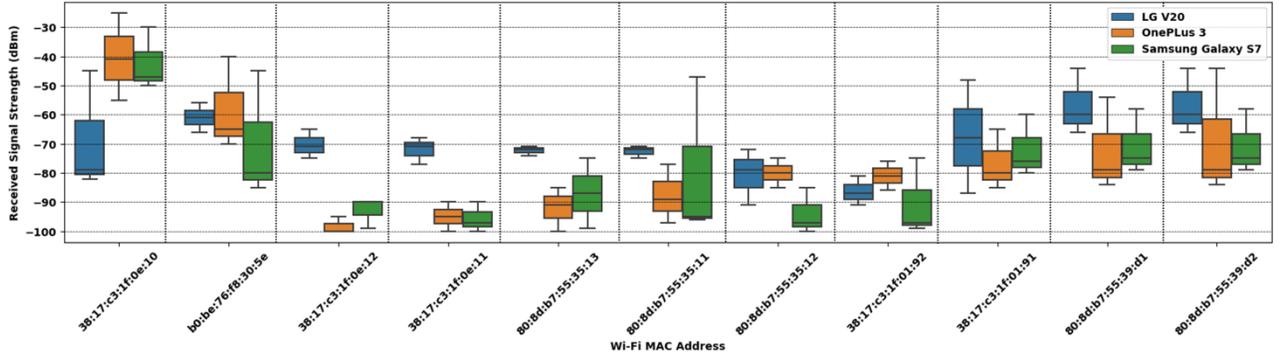

Figure 1: Analysis of device heterogeneity, across different smartphone at the same time and location

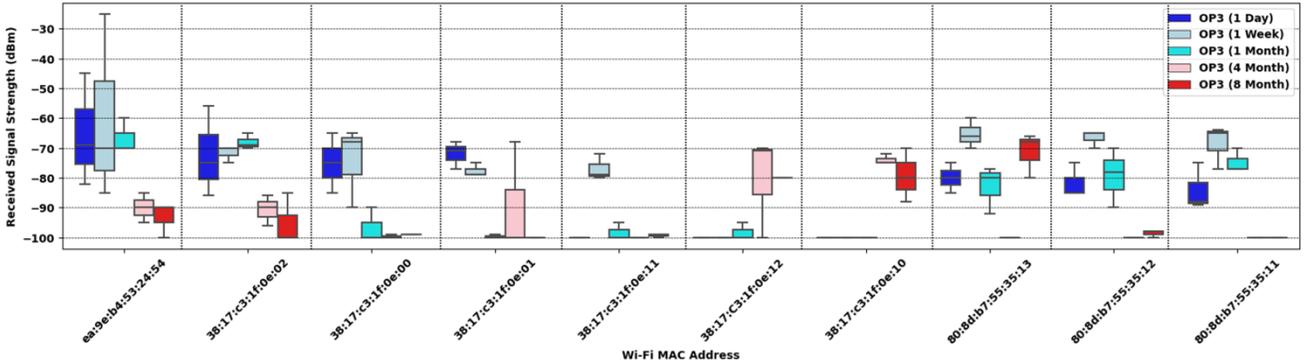

Figure 2: Analysis of temporal variations across varying time periods as captured by the same smartphone at the same location.

Temporal variations play a crucial role in the real-world adoption of any indoor localization system, as they cause RSS fingerprints to undergo significant changes over time, making indoor localization increasingly complex. To analyze the effects of temporal variations, we conducted Wi-Fi RSS data collection instances (CI) over several months at the same location. For each CI, we captured multiple fingerprints and subsequently averaged them to derive representative results, Figure 2 shows the results of the captured Wi-Fi RSS data with a OnePlus 3 smartphone. The data was collected during normal working hours to incorporate external noise, in addition to temporal variations caused by changes with the Wi-Fi APs.

One can observe significant fluctuations in RSS values for APs over the course of several months. For instance, on day 1, the average RSS value for the AP with MAC ID '38:17:c3:1f:0e:02' was -75 dBm. However, at week 1 and at month 1 we observed changes for the RSS value for the AP to -70 dBm and -69 dBm (on average), respectively. Although these changes may not be considered significant, a more substantial variation occurred at months 4 and 8 when the RSS value dropped significantly to -90 dBm and -100 dBm (on average), respectively. These fluctuations in RSS values can be attributed to the impact of temporal variations, encompassing changes in AP configurations, both short-term and long-term variations. Such variations can have a detrimental impact on the performance of ML models used for localization, potentially resulting in mispredictions. We also noticed similar variations for other APs over the months. For example, considering the AP with MAC ID '38:17:c3:1f:0e:10', we observed an average RSS value of -80 dBm at day 1. However, at week 1 and at month 1, the RSS value dramatically dropped to -100 dBm (on average), indicating that the AP was turned off during that period. Interestingly, at months 4 and 8, we observed a reversal of the previous state, as the AP became active again with RSS values of -70 dBm and -80 dBm (on average), respectively. These observations highlight the dynamic nature of indoor environments and the impact of temporal variations on RSS readings. Addressing these challenges is crucial for developing a robust and long-term indoor localization solution.

## V. STELLAR FRAMEWORK

The STELLAR framework, an overview of which is shown in Figure 3, aims to address indoor localization challenges through a two-phase process. In the offline phase (red arrows), RSS fingerprints are collected from various reference points (RPs) across the building's floorplan. Each row in the training database represents the RSS values of visible APs for a specific RP, forming a single fingerprint. These fingerprints are then organized into three sets (triplets): anchor, positive, and negative, and sent to a Siamese multi-headed attention neural network model for training. The underlying attention model uses the inputs from each triplet to generate embedding hyper-spaces for each of the triplets, capturing characteristics related to device heterogeneity and temporal variations. Details related

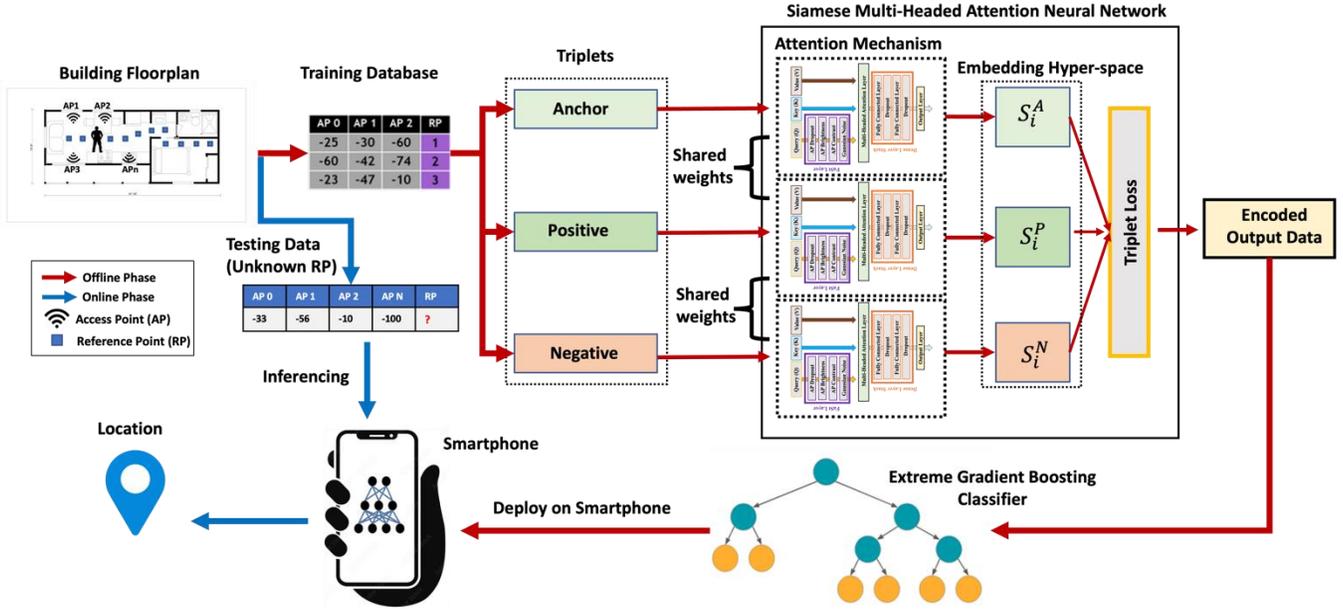

Figure 3: Working of the STELLAR framework, including offline (training) phase and online (inferencing) phase.

to triplet selection and attention mechanism are discussed in the later sub-sections.

The triplet loss trains the multi-headed attention model's output (the three embedding hyper-spaces), resulting in encoded versions of the triplets. The encoded data, along with associated RP locations, is then used to train an extreme gradient boosting (XgBoost) classifier. The XgBoost classifier improves performance and compensates for any negative impact from sub-optimal triplets. After this step, the offline phase concludes, and the trained ML models are now deployed on the smartphone for the online phase to predict location.

In the online phase, a user captures Wi-Fi RSS from different APs, using a smartphone at an unknown location. The RSS values are used to create a Wi-Fi fingerprint. This fingerprint is inferred by the Siamese model, generating an encoded output. The XgBoost classifier then processes this encoded data to predict the user's location.

The two-phase approach enables the STELLAR framework to address indoor localization complexities, incorporating device heterogeneity and temporal variations. Unlike most ML-based methods that typically require extensive training data collected from various locations, STELLAR is designed to deliver robust indoor localization with minimal training data. This efficiency is facilitated by the novel Siamese multi-headed attention neural network, combined with the proposed contrastive learning technique. These components empower STELLAR to achieve superior localization performance, even when faced with extreme RSS variations caused by temporal changes, all while using a limited amount of training data. In the following sub-sections, we describe the components of the STELLAR framework in more detail. In the following sub-sections, we describe the components of the STELLAR framework in more detail.

### A. RSS Fingerprint Pre-Processing

The database shown in Figure 3 contains RSS values for different Wi-Fi APs along with their corresponding RPs. The RSS values range from -100dBm to 0dBm, with -100 indicating no signal and 0 indicating the strongest signal, as discussed earlier. To prepare the training data for the Siamese multi-headed attention model (or inference data for prediction), the captured RSS values are normalized to a range of 0 to 1, where 0 represents weak or null signals, and 1 represents the strongest signal. Each fingerprint in this dataset is then used as a triplet input for the Siamese model. The details on selecting the three triplets are presented next.

### B. Triplet Selection for Contrastive Learning

After pre-processing the fingerprints from the training database, they are divided into three triplets: *anchor*, *positive*, and *negative*, as shown in Figure 3. These triplets serve as inputs to train the Siamese multi-headed attention neural network. Each triplet is carefully selected to allow efficient training of the Siamese model.

The *anchor* consists of the captured fingerprint at a particular RP for training. The *positive* is formed by using the same fingerprint as the anchor, but with changes to a subset of RSS values. The *positive* is formed by randomly replacing a subset of the RSS values to -100dBm or 0 (normalized) to simulate changes in APs that can arise in the indoor space. For instance, AP:0 can be turned off due to maintenance or power outage and this can change the captured RSS fingerprint leading to inaccurate localization, simulating random AP dropout in the *positive* can prepare the model to address such variations. The *negative* is another fingerprint from the training database but located at a different RP from the anchor. For instance, if the *anchor* fingerprint is collected at RP:1, the *negative* fingerprint can be collected from an adjacent RP, such as RP:0 or RP:2.

Our Siamese model utilizes contrastive learning [33] during training to learn to identify similarities between the *anchor* and *positive* triplets, enabling it to adapt to scenarios where input RSS fingerprints are similar to the *anchor* but may have variations. Similarly, the model learns to differentiate between the *anchor* and *negative* triplets, allowing it to discern fingerprints from different RP locations, even when the *anchor* and *negative* triplets are close in locations. This powerful learning method finds similarities and dissimilarities between fingerprints, even when they exhibit low or high similarity characteristics, respectively. The Siamese model outputs an encoded representation of the triplets used for training.

*B.1 Anchor Triplet Selection*

The anchor represents the captured input fingerprint data used for training the Siamese model. It consists of RSS fingerprints captured at various RPs within the indoor environment, each having a unique set of visible Wi-Fi APs with their corresponding RSS values. The anchor plays a critical role as it serves as the baseline for training the network. It acts as a baseline against which the model learns to identify similarities and differences with other triplets (positive and negative). By using the anchor triplet as the basis for comparison, the Siamese model can effectively capture essential patterns and relationships within the data, enabling accurate localization predictions during the online phase.

*B.2 Positive Triplet Selection*

The positive triplet is an essential part of the training process in the STELLAR framework. It is derived from the anchor triplet, which represents the input fingerprint data used to train the Siamese model. The positive triplet is created by randomly selecting a certain percentage (D%) of APs from the anchor triplet and setting their corresponding RSS values to -100 dBm (or 0 after normalization), effectively simulating the dropout of these APs. The introduction of dropout in the positive triplet serves the purpose of simulating the temporal variations that commonly occur in real-world indoor environments. By training the Siamese model using positive triplets with AP dropouts, the model learns to handle missing APs in the input data, improving its resilience to temporal variations.

*B.3 Negative Triplet Selection*

The purpose of introducing this negative triplet is to provide examples of fingerprints that closely resemble the anchor triplet, but at a different RP. This helps the Siamese model to learn to distinguish between similar yet distinct RPs and their corresponding fingerprints, enabling more accurate localization.

To find the negative triplet that shares the closest similarity with the anchor triplet, we use the Euclidean distance (ED) method, as shown in Eq. (1). The ED method calculates the squared differences between the RSS values of the APs in the anchor and negative triplets, sums them, and takes the square root to obtain the final ED value. By measuring the overall differences between the triplets, the ED metric allows the model to evaluate their relative similarity or dissimilarity.

$$ED(F_{Anchor}, F_{Neg}[i]) = \sqrt{\sum (F_{Anchor} - F_{Neg}[i])^2} \quad (1)$$

Eq. (1) represents the calculation of the ED between the anchor triplet ($F_{Anchor}$) and the negative triplet ($F_{Neg}$). The variable *i* refers to the RP location different from the anchor's RP. The selection process of the negative fingerprint ($F_{Neg}$) is shown in Eq. (2).

$$F_{Neg} = Argmin\ (ED(F_{Anchor}, F_{Neg})). \quad (2)$$

The framework uses the $Argmin$ operation to find the fingerprint with the minimum distance to the anchor triplet ($F_{Anchor}$). In other words, ($F_{Neg}$) (negative triplet) is chosen as the fingerprint from the training database that exhibits the closest similarity to the anchor fingerprint according to the ED metric. This process forces the Siamese model to learn dissimilarities between the similar triplets, as they are from different RPs.

### C. Siamese Multi-Headed Attention Neural Network

Our Siamese model comprises of three multi-headed attention neural networks, each trained to accept one of the three triplets (anchor, positive, and negative) for training. These attention mechanisms share weights, enabling the model to learn common patterns and relationships across the triplets, enhancing its ability to handle challenges such as temporal variations and device heterogeneity effectively. During training, the multi-headed attention neural network generates embedding hyper-spaces for each triplet, capturing essential features and relationships within the fingerprint data (from the triplets), as shown in figure 3. Utilizing the triplet loss with contrastive learning, the model learns to identify similarities between the anchor and positive triplets and distinguish between the anchor and negative triplet, providing a comprehensive understanding of the fingerprints' characteristics, as discussed in Section V.B. The model's final output is an encoded representation of the three triplets used for training, containing valuable information about potential temporal characteristics and device heterogeneity. By leveraging the multi-headed attention neural network and the triplet loss with contrastive learning, the Siamese model provides insights into the indoor environment's complexities, making it more resilient to temporal changes and heterogeneity, thereby leading to improved localization performance.

*C.1 Multi-Headed Attention Neural Network*

The multi-headed attention neural network within the Siamese model is designed with three inputs: Query (Q), Keys (K), and Values (V), as illustrated in figure 4. Each of the three multi-headed attention neural networks handles one of the three triplets, with the triplet data fed into the Q input. Before passing through the Q input, the triplet data undergoes domain-specific augmentation using the FaSt (Fingerprint Augmentation Stack) module, discussed in Section V.C.2. The K input represents the entire training database (fingerprints across all RPs), enabling comparisons with the Q input. On the other hand, the V input signifies the corresponding RP locations, of all the fingerprints in K in a one-hot encoded form. This one-hot encoding technique effectively represents categorical variables as binary vectors, where only one element is set to 1 while the rest are set to 0, enabling the network to process categorical information efficiently. This multi-headed attention mechanism allows the Siamese model to capture and focus on critical patterns and relationships within the fingerprint data for accurate indoor localization.

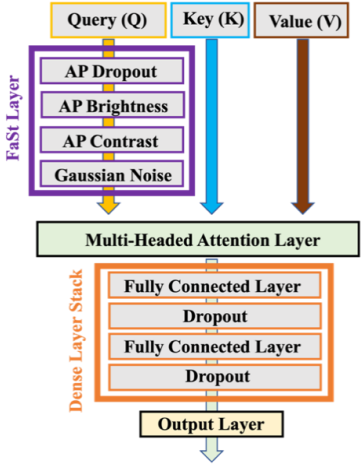

Figure 4: Multi-headed attention mechanism

The Q, K, and V are then fed into the multi-headed attention layer, as shown in figure 4. The attention layer works by computing attention weights for each pair of Q and K elements and applying these weights to the corresponding V elements. The attention weights are calculated using the scaled dot-product attention mechanism, which measures the similarity between the Q and K elements. The attention layer allows the network to focus on relevant features in the fingerprint data and the corresponding RP locations, facilitating accurate indoor localization.

The attention calculation is represented as follows:

$$Attention(Q,K,V) = softmax\left(\frac{QK^T}{\sqrt{d_k}}\right)V \quad (3)$$

In the context of Wi-Fi RSS fingerprinting based indoor localization, Q represents the input fingerprint, K represents the fingerprint database (fingerprints across all RPs), and V represents the corresponding RP locations. The dot product between Q and K captures the similarity between the input fingerprint and the fingerprint database, while the $softmax$ function normalizes the attention weights based on their relative strengths. The resulting attention weights are then applied to V to obtain the attended representation.

In multi-headed attention, this attention mechanism is applied multiple times with different linear projections of the Q, K, and V matrices. Let $h$ represent the number of attention heads. The multi-headed attention can be mathematically represented as:

$$MultiHead(Q,K,V) = Concat(h_1, h_2, \dots h_n)W^o$$
$$where\ h_i = Attention(QW_i^Q, KW_i^K, VW_i^V) \quad (4)$$

After the attention layer, the output is passed through a Dense Layer Stack, which consists of a series of fully connected layers with dropout regularization. The Dense Layer Stack effectively learns and extracts meaningful representations from the output of the attention layer, as shown in Figure 4. These layers enable the network to capture complex relationships and patterns in the data while preventing overfitting.

The multi-headed attention layer in the Siamese model uses shared weights across all heads to enable the layer to process all three triplets (anchor, positive, and negative) simultaneously. The resulting outputs for each triplet are stored as embedding hyper-spaces (*i*). These embedding hyper-spaces, representing the anchor ($S_i^A$), positive ($S_i^P$), and negative ($S_i^N$), are then trained using triplet loss (discussed in Section V.C.3). The combination of the attention mechanism and the Siamese method allows addressing the challenges posed by both heterogeneity and temporal variations during indoor localization.

### C.2 Fingerprint Augmentation Stack (FaSt Layer)

The FaSt module incorporates various powerful features aimed at enhancing the accuracy and robustness of indoor localization in real-world environments. One significant component is random AP dropout, wherein specific APs are randomly dropped from the RSS fingerprint data, followed by infilling the missing APs using Gaussian noise. This strategic approach can be effective in addressing device heterogeneity. As the model becomes more adept at handling diverse indoor environments with varying signal strengths.

In addition to random AP dropout and Gaussian noise, the FaSt module introduces two essential components: AP random brightness and AP random contrast. These features introduce variations in brightness and contrast levels, respectively, into the RSS fingerprint data. Brightness signifies the intensity or strength of the signal received from the APs, while contrast measures the difference between the highest and lowest signal strengths in the data. The module alters these signal strengths, by altering the intensities for AP random brightness and scaling the difference between the highest and lowest signal strengths for AP random contrast, effectively simulating different levels of signal intensity and contrast variations in the dataset. By incorporating these diverse patterns, the FaSt module equips the ML model with a comprehensive understanding of the potential variations that may be caused due to device heterogeneity.

### C.3 Triplet Loss

The triplet loss method plays a crucial role in the Siamese model, enabling it to learn similarities and dissimilarities between the triplets through contrastive learning. This loss function operates on the embedding hyper-spaces generated from the three triplets: anchor, positive, and negative. The relative distances between these embeddings in the hyper-space are used to optimize the overall model.

The main objective of minimizing the triplet loss is to ensure that the anchor and positive triplets, which represent similar fingerprints are close to each other in the hyper-space. At the same time, the anchor and negative triplets, which correspond to fingerprints from different RPs, should be pushed farther apart. This process facilitates the Siamese model in effectively identifying similarities between altered fingerprints (due to temporal variations or device heterogeneity) and distinguishing between fingerprints at different RPs, enhancing the accuracy and resilience of the indoor localization framework. By leveraging the triplet loss and contrastive learning, the Siamese model can create meaningful embeddings that are stored as the encoded output data. The mathematical equation for the triplet loss are:

$$F\ (Pos) = Argmin\ (||S_i^A - S_i^P||)^2 \quad (5)$$
$$F(Neg) = Argmax\ (||S_i^A - S_i^N||)^2 \quad (6)$$

$$Triplet\ Loss = \sum(F(Pos) - F(Neg)) \quad (7)$$

where $S_i^A$ is the embedding hyper-space representation of the anchor triplet, $S_i^P$ is the embedding hyper-space representation of the positive triplet, and $S_i^N$ is the embedding hyper-space representation of the negative triplet. The triplet loss is used to output the encoded versions of the triplets.

### D. Extreme Gradient Boosting (XgBoost)

The encoded output from the Siamese multi-headed attention neural network is fed into an Extreme Gradient Boosting (XgBoost) classifier for location prediction. As an ensemble learning algorithm, XgBoost learns from the encoded data and maps the learned features to their corresponding RP locations, enabling precise and robust indoor localization predictions. XgBoost combines the predictions of multiple weak models, called decision trees, to make a final prediction. Each decision tree is built iteratively, with each subsequent tree aiming to improve upon the previous ones. During training, the XgBoost algorithm computes a loss function (multi-class loss), which quantifies the difference between the predicted and target values. The algorithm then determines the optimal structure of the decision tree by minimizing the loss function.

Mathematically, the XgBoost algorithm aims to learn a prediction function $F(x)$ that maps the encoded data to the target location coordinates. The prediction function $F(x)$ can be represented as:

$$F(x) = W_0 + \sum W_j * F_j(x) \quad (8)$$

where $W_0$ is the bias term, $W_j$ is the weight associated with the *j-th* decision tree, and $F_j(x)$ is the prediction made by the *j-th* decision tree for the input *x*. During training, the XgBoost algorithm optimizes the weights ($W_j$) and bias term ($W_0$) by minimizing a combination of the loss function. This process ensures that the decision trees collectively capture the complex relationships between the encoded input data and the target location coordinates.

## VI. EXPERIMENTS

### A. Experimental Setup

In this section, we describe the experimental setup to assess the performance of our proposed STELLAR framework in real-world scenarios and compare it with state-of-the-art indoor localization frameworks, including: the multi-headed attention framework (ANVIL) [15], the vision-transformer framework (VITAL) [26], the autoencoder support gaussian process classifier framework (WiDeep) [28], the stacked autoencoder with gradient boosting framework (SANGRIA) [29], and the long-term K-nearest neighbor framework (LT-KNN) [30]. Our experiments test the frameworks for resilience towards both device heterogeneity and temporal variations. To ensure resilience of all frameworks towards these challenges, we train all the frameworks using data collected from a single smartphone across multiple RPs in the building floorplan at the same collection instances (CI) or time period. This training enables the frameworks to be evaluated on various smartphones over an extended period without requiring re-training.

The training process of the STELLAR framework begins by setting the D% of APs (in the negative triplet) to be dropped to 60% (see Section V.B.3) and then training the Siamese multi-headed attention neural network with 7 heads and a head size of 50, where each head represents an individual sub-network module responsible for processing input features and head size specifies the dimensionality of each head (see Section V.C.1). The FaSt module, consisting of AP dropout (0.1), AP random contrast (0.1), AP random brightness (0.1), and Gaussian noise (0.12), all designed to introduce variability that can arise due to heterogeneity and enhance the robustness and adaptability of the model (see Section V.C.2). The Dense Stack Layer, comprising fully connected layers with ReLU (rectified linear unit) nonlinearity, has 128 and 64 neurons, respectively, while the output layer uses the Softmax function (see Section V.C.1). During training, the Siamese multi-headed attention neural network is optimized with a learning rate of 0.0001 for 300 epochs and is trained using the triplet loss (see Section V.C.3), resulting in a total of 201,672 trainable parameters. The XgBoost classifier is set with a maximum depth of 7 (see Section V.D).

To evaluate the performance of a localization framework, we measure the localization error using the Euclidean distance (ED), which quantifies the dissimilarity between the predicted and actual (target) location. This metric provides valuable insights into the effectiveness of a framework in accurately estimating positions in indoor environments. The localization error is calculated based on the distance between the ground truth location and the estimated location of a device, which is based on the RSS data collected by the device. This approach allows us to assess the localization error of the framework across multiple devices, building scenarios, and time periods, ensuring that the framework can be reliably used in real-world settings.

### B. Data Collection and Pre-Processing

To gather information for our indoor localization experiments, we utilized six smartphones from different manufacturers, each equipped with a distinct Wi-Fi chipset, as shown in Table 1.

Table 1: Details of smartphones used in our experiments.

| Manufacturer | Model | Acronym | Wi-Fi Chipset |
|---|---|---|---|
| BLU | Vivo 8 | BLU | MediaTek Helio P10 |
| HTC | U11 | HTC | Qualcomm Snapdragon 835 |
| Samsung | Galaxy S7 | S7 | Qualcomm Snapdragon 820 |
| LG | V20 | LG | Qualcomm Snapdragon 820 |
| Motorola | Z2 | MOTO | Qualcomm Snapdragon 835 |
| OnePlus | 3 | OP3 | Qualcomm Snapdragon 820 |

It is worth noting that although some of the Wi-Fi chipsets used are the same, each smartphone manufacturer implements different firmware for noise filtering, resulting in inherent heterogeneity even when the Wi-Fi chipsets are identical. We selected these devices to capture both heterogeneity due to hardware and software.

We conducted experiments in two distinct buildings, each with unique characteristics, to evaluate the indoor localization frameworks, as shown in Figure 5. Building 1 featured a combination of wood and cement materials, spacious areas with computers, and a total of 160 identifiable Wi-Fi APs. Building 2 had a modern design with a mix of metal and wooden structures, open spaces, and shelves equipped with heavy metallic hardware. We identified 218 unique APs in Building 2. These buildings were located close to each other, separated by a single wall, enabling us to study the impact of interference from different sources across the buildings.

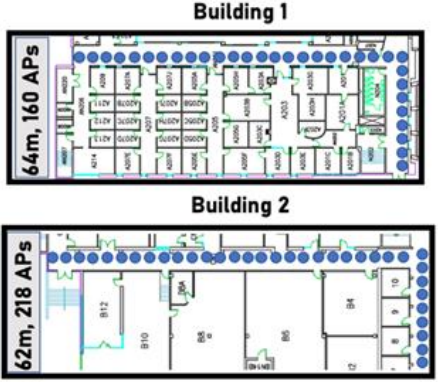

Figure 5: Floorplan layout for Building 1 (64 meters long with 64 RPs) and Building 2 (62 meters long with 62 RPs) and RP (blue dots).

The complex indoor environment, coupled with the presence of metal and electronic devices near the test paths, resulted in noisy Wi-Fi fingerprints, posing challenges for accurate indoor localization. To capture realistic scenarios, we collected Wi-Fi RSS fingerprints during regular working hours, without artificially controlling human occupancy. We collected a total of 6 fingerprints per RP at each collection instance (CI), spanning a duration from 30 seconds to several months. We captured 16 CIs (CIs:0-15) over a total span of 8 months. To capture the impact of varying human activity throughout the day, we performed the first three CIs (0-2) for both paths on the same day, with a time interval of 6 hours between each CI. This allowed us to capture fingerprints early in the morning (8 A.M.), at mid-day (3 P.M.), and late at night (9 P.M). Subsequently, we performed CIs 3 to 8 across 6 consecutive days. The remaining CIs (9-15) were carried out at approximately 30-day intervals.

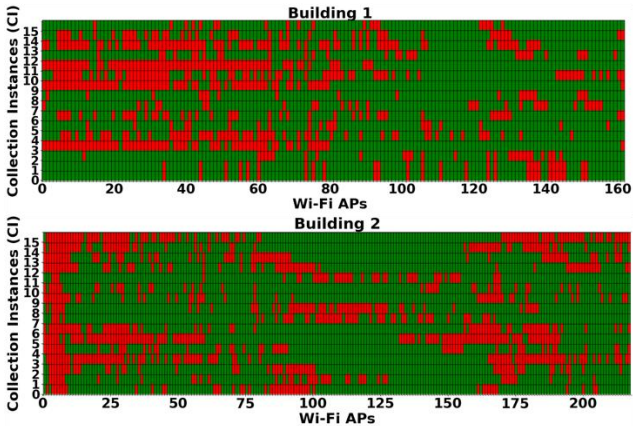

Figure 6: Analyzing visible (green) APs and invisible (red) APs over time.

Figure 6 illustrates the ephemerality of Wi-Fi APs along the Building 1 and Building 2 paths across the 16 CIs (CIs:0-15) over the total duration of 8 months. Each CI is represented on the y-axis, while the x-axis displays the specific Wi-Fi APs. Red marks indicate that a particular AP was not observed during the corresponding CI and green marks indicate visibility of the AP. We did not observe any significant changes in AP visibility up to CI:2. However, between CI:3 and CI:9, approximately 20% of the Wi-Fi APs became unavailable. After CI:10 approximately 40% of the Wi-Fi APs became unavailable. For the Building 1 and Building 2 paths, we specifically utilized a subset of CI:0, capturing fingerprints during the early morning, for the offline phase of our experiments. This means that the training process solely relied on this subset of data from CI:0. The remaining data from CI:0 and CIs:1-15 were reserved for testing purposes.

By collecting Wi-Fi RSS fingerprints over a duration of seconds to months, including customary working hours, we were able to capture various forms of interference and obstacles that impact Wi-Fi signal propagation. This approach allowed us to account for dynamic factors such as population density and human movement within the building, providing a realistic understanding of the challenges encountered in real-world indoor localization scenarios. In our dataset, we split the data in a 5:1 ratio to ensure a robust evaluation of the model's performance. During the training phase, we utilized 5 fingerprints per RP, while for the online phase, we used 1 fingerprint per RP. We employed a granularity of 1-meter separation between RPs, which we deemed sufficient for accurate localization of individuals. *Additionally, we have made our data openly available [34] for reproducibility and to support further research in the indoor localization community.*

### C. Experimental Results

In this section, we present the results of various experiments conducted to assess the performance of our STELLAR framework. We first present the outcomes of hyperparameter tuning to optimize the STELLAR framework. Subsequently, we conduct an ablation study to explore the selection process of the gradient boosting classifier. Next, we compare the STELLAR framework with state-of-the-art indoor localization frameworks to evaluate its effectiveness in handling device heterogeneity. Additionally, we compare the STELLAR framework with existing approaches to assess its performance under temporal variations. Lastly, we conclude by subjecting all frameworks to an extended evaluation, considering newer diverse devices (different devices from table 1) and extreme temporal variations to ascertain their robustness under extremely challenging conditions.

#### C.1 Hyperparameter tuning: AP Dropout

In this experiment, we investigate the impact of dropping a percentage (D%) of APs (for the negative triplet, see section V.B.3) during the training of the Siamese multi-headed attention neural network in the STELLAR framework. In this experiment, we investigate the impact of dropping a percentage (D%) of APs (for the negative triplet, see section V.B.3) during the training of the Siamese multi-headed attention neural network in the STELLAR framework. We randomly drop D% of APs to introduces variability, enabling the model to effectively adapt to variations arising from temporal changes. This randomization simulates scenarios where APs may be missing or turned off without following a predefined pattern. The purpose of this experiment is to analyze how the framework performs when dealing with scenarios where certain APs may be missing or turned off. For the experiment, we selected the OP3 device as the training device and used a subset of data from CI:0 for training the STELLAR framework.

The performance of the framework was evaluated on the remaining devices across all CI values (ranging from 0 to 15) across both the building floorplans. The localization error was reported in meters and calculated using ED, without re-training on any of the CIs (re-calibration-free). We varied the value of D% (at the negative triplet, see section V.B.3) from 10% to 90% in steps of 10%, representing the percentage of APs that are dropped during training. The dropped APs were simulated by setting their corresponding RSS values to -100 dBm (or 0 after normalization). The average localization error per value of D% was computed by considering the localization errors from all testing devices and CIs. These results are shown in Figure 7.

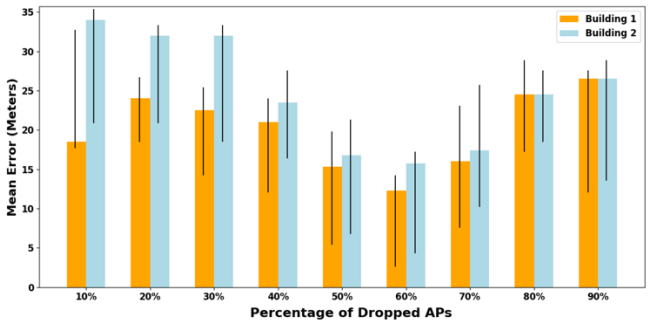

Figure 7: Tuning D% of AP dropout during offline phase for both building floorplans with min-max intervals showing best and worst-case errors for each D% value.

From the results, we can observe that as the percentage of dropped APs increases from 10% to 60%, the average localization error decreases gradually. This suggests that the framework can adapt and compensate for the missing APs. However, beyond 60% of dropped APs, the average localization error starts to increase. This behavior may occur because a higher percentage of dropped APs introduces a greater loss of information, making it more challenging for the framework to accurately estimate the device's location. Thus, highlighting the importance of selecting an appropriate value of D to ensure accurate indoor localization. By identifying the optimal value of D, we can enhance the framework's performance and ensure its robustness in real-world scenarios. In this study, we use the specific value of D (60%) that demonstrated the best performance for all the subsequent experiments.

### C.2 Ablation study: Post Encoding Classifiers

In this ablation study, we explore the performance of different non-parametric models when training the output from the Siamese multi-headed attention neural network (encoded output). The preference for non-parametric models arises from their capability to handle the complexities of the encoded output without the constraints of fixed parameters, resulting in more flexible and adaptive solutions. The non-parametric model is essential to prevent suboptimal triplet selection which can hinder learning and negatively impact the model's performance (see section V.C.3). The goal is to compare the accuracies achieved by popular non-parametric algorithms: KNN (k-nearest neighbors) where the number of neighbors is set to 4 for determining the nearest data points for prediction, RF (Random Forest) with 100 estimators for creating an ensemble of decision trees, SVM (Support Vector Machine) with the radial basis function (RBF) as a kernel to transform the data and find optimal decision boundaries, CatBoost (categorical gradient boosting) with a depth of tree set to 7, and XgBoost (extreme gradient boosting) with a depth of tree to 7, all chosen based on the values that provided the best results during the exploration process.

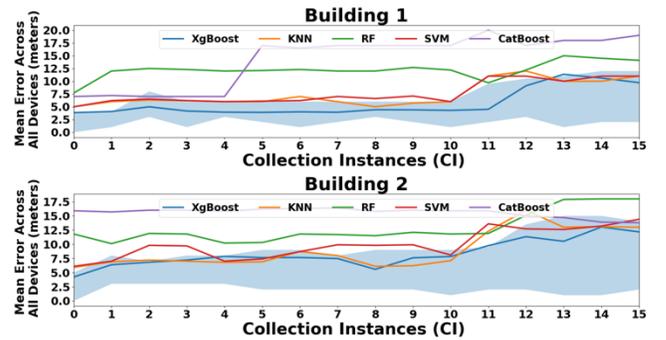

Figure 8: Comparing response of different classifier in temporal space for both building floorplans. Blue shadow indicates best and worst-case errors for STELLAR.

We selected the OP3 device as the training device and utilize a subset of data from CI:0 to train all frameworks. The performance of the framework is then evaluated on the remaining devices (from table 1) across all CI values (ranging from 0 to 15) across both the building floorplans. The localization errors, reported in meters, are averaged across all devices for each CI, and are shown in Figure 8.

The KNN, SVM and XgBoost models consistently outperform the other models, particularly at CI greater than 9 where a significant number of APs are dropped. The KNN model excels in capturing local patterns and finding nearest neighbors, resulting in low localization errors. The SVM model performs well due to its ability to create effective decision boundaries and handle complex data distributions. The XgBoost excels as the most effective model, employing a powerful ensemble of decision trees to capture intricate data relationships, robustly handling complexities from device heterogeneity and temporal variations, resulting in superior indoor localization accuracy. RF exhibits stable performance but is less adept at handling the dropped AP scenario, as it relies on averaging the predictions of multiple decision trees, and each tree is constructed independently using a random subset of features and data points. CatBoost, a gradient boosting algorithm designed to handle categorical features, may perform poorly in scenarios involving dropped APs, exhibiting similar limitations to the RF algorithm. As a result, it may not be optimal for addressing the challenges presented in this study. In conclusion, based on the analysis, XgBoost proves to be the top performer demonstrating the lowest localization error, adaptability to handle heterogeneity, and temporal variations, making it the ideal choice for subsequent experiments in long-term indoor localization.

### C.3 Ablation study: Sensitivity of training data

In this ablation study, we delve into the sensitivity of the STELLAR framework to variations in the training data. Our objective is to assess the framework's performance across different smartphones and building floorplans by adjusting the number of fingerprint samples collected per RP, during the training phase. We explore various training scenarios, ranging from utilizing 1 to 5 fingerprint samples per RP, as shown in Figure 9. The X-axis represents all CIs over a total span of 8 months, while the Y-axis denotes the mean localization errors (averaged across all smartphones and RPs within the respective building floorplans). Additionally, we include a shaded region for the best performing sample to display the variability of localization performance (with the region spanning between the best and worst-case scenarios). We selected OP3 as the training device from CI:0 for all training samples, , this choice was made to demonstrate the robustness of the STELLAR framework, even when trained with a single or limited training device. This choice underscores the independence of the STELLAR framework from the choice of the training device. Further details can be found in Section VI.C.5, which depicts the minimal impact on STELLAR's performance when utilizing different training devices. The designation CI:0 indicates that the training data used was collected only from the first instance.

Our findings indicate that the fingerprint samples collected per RP during the training phase impacts the efficacy of STELLAR. Notably, employing just one fingerprint per RP results in the poorest performance. This can be attributed to the inadequacy of the training data to account for the multifaceted challenges posed by indoor environments, such as device heterogeneity, temporal variations, and the presence of external noise. Conversely, as the number of fingerprint samples per RP increases from two to five, STELLAR's performance exhibits a notable and consistent improvement. This is reflected in the reduction of localization errors across both temporal and heterogeneity spaces.

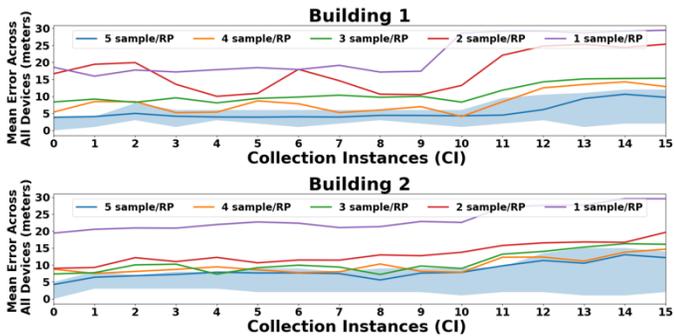

Figure 9: Sensitivity analysis of training data on STELLAR's performance. Blue shadow indicates best and worst-case errors for STELLAR.

### C.4 Comparison vs. state-of-the-art (Heterogeneity)

In this experiment, we evaluate the performance of the STELLAR framework in comparison to state-of-the-art localization frameworks, with a specific focus on device heterogeneity. To ensure transparency and reproducibility, we specifically selected the state-of-the-art localization frameworks with open-source implementations. To assess the impact of different training devices, we train each framework using data from a single device and evaluate its accuracy on the remaining testing devices. By observing any significant changes in performance across different training devices, we can determine the framework's resilience towards heterogeneity.

Figure 10 presents the results of this analysis, showcasing the performance of all frameworks in a matrix format while Figure 11 summarizes the average errors. In Figure 10, the x-axis represents the training devices, while the y-axis represents different frameworks, and the z-axis represents the average localization errors (meters) across all the testing devices. The experiments were performed on both the floorplans (building 1 and 2). We specifically selected the CI:0 data from both training and testing devices in the building 1 and building 2 floorplans. All frameworks aim to address device heterogeneity except LT-KNN which is designed to address temporal variations.

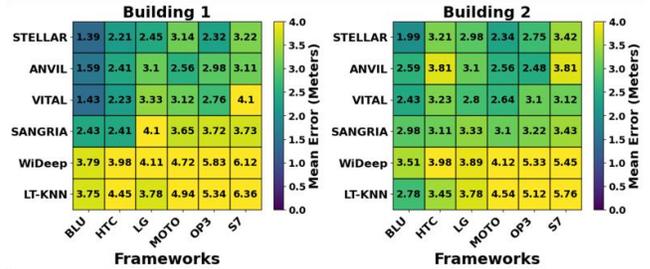

Figure 10: Framework comparison: average localization errors (meters) across devices.

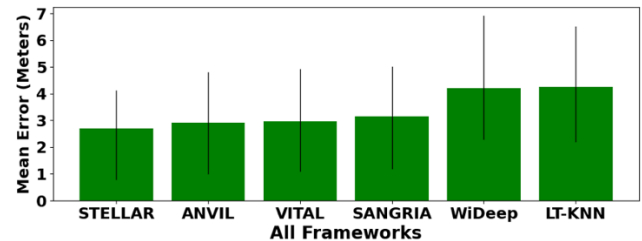

Figure 11: Average errors across training - testing devices from both building floorplans with min-max intervals showing best and worst-case errors for each framework.

As an example of how to interpret Figure 10, the first column for both buildings are the case where the BLU device was used as a training device, and the numbers in the column are the average localization error across the testing devices (remaining 5 smartphones from Table 1). It can be observed that ANVIL performs well due to its attention-based approach that captures relevant spatial dependencies. VITAL also excels with its vision transformer-based model, effectively modeling complex relationships. SANGRIA's stacked autoencoder architecture enables robust handling of device heterogeneity.

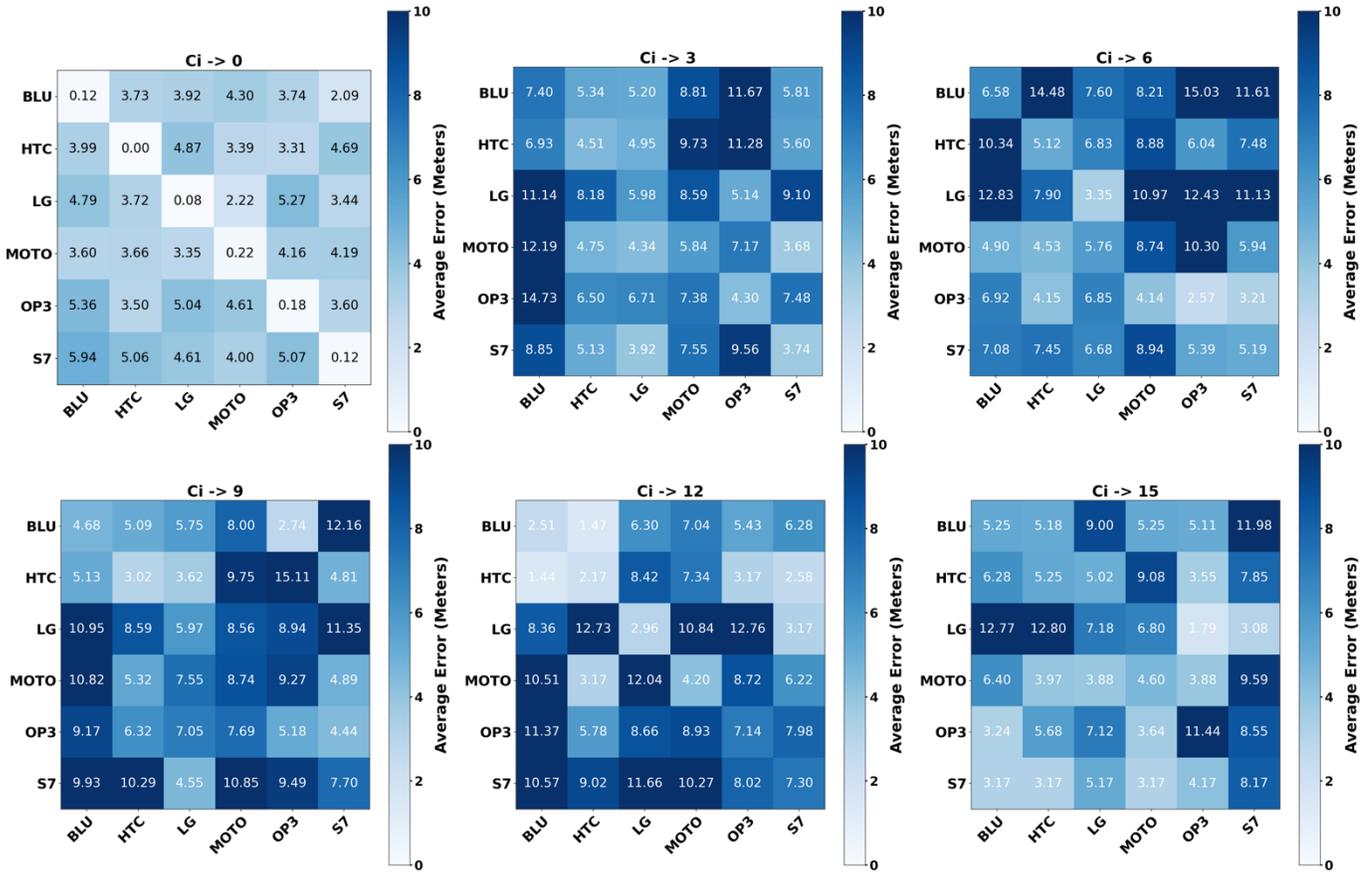

Figure 12: STELLAR's performance analysis: average localization error (meters) across training and testing devices with varied temporal variations (CIs)

STELLAR reports the lowest localization errors compared to its competitors. From Figure 11, STELLAR shows improvements over ANVIL (by 8.81%), VITAL (by 9.19%), SANGRIA (by 24.9%), WiDeep (by 72.41%), and LT-KNN (by 75%) in addressing device heterogeneity.

### C.5 Comparison vs. state-of-the-art (Temporal Variations)

In this study, we analyze the performance of the STELLAR framework in terms of temporal variations and compare it with state-of-the-art localization frameworks. We also conduct an in-depth analysis of the framework's ability to address heterogeneity in the temporal space. Figure 12 illustrates the performance of STELLAR across all training and testing devices, and across six different CIs. The x-axis represents the testing device, the y-axis represents the training device, and the z-axis represents the average localization error in meters. We focus on specific CIs: CI:0 (day 1), CI:3 (1 week), CI:6 (1 month), CI:9 (2 months), CI:12 (5 months), and CI:15 (8 months) to explore performance with temporal variations across different timespans. For brevity, we omit the results of other CIs, as they exhibit similar patterns to those depicted in Figure 12.

To further assess STELLAR's performance across different CIs, we present Figure 13, which displays the variance between the best and worst-performing testing devices for specific training devices and CIs. The x-axis represents the training device, the y-axis denotes the CIs, and the z-axis indicates the difference in performance between the best and worst testing devices. Figure 13 analysis reveals that, for the majority of cases, STELLAR exhibits consistent error rates, even in the presence of temporal variations. This indicates the ability of the STELLAR framework to address the challenges of heterogeneity, primarily attributed to its attention mechanism that effectively handles the fluctuating nature of RSS values. Furthermore, we note that STELLAR maintains strong performance even at later CIs (greater than CI:9), demonstrating its ability to handle temporal variations. This robustness can be attributed to the contrastive learning, which effectively captures variations in the temporal space and enables STELLAR to serve as a long-term, resilient localization framework.

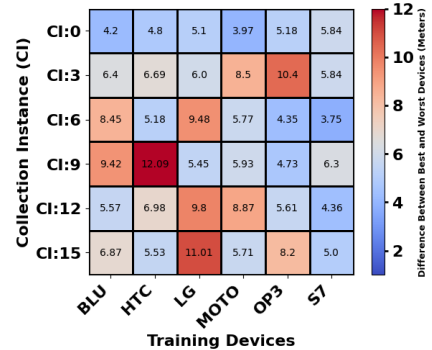

Figure 13: Difference between best and worst performing testing device per training device per CI.

To gain deeper insights into the performance of the STELLAR framework, we present Figure 14, showcasing the average localization errors across different CIs in a line plot. This analysis includes a comparison with state-of-the-art localization frameworks, focusing on the responses of each framework in different building floorplans, capturing both temporal and heterogeneity aspects. From Figure 14, it becomes evident that STELLAR consistently achieves lower localization errors compared to other frameworks, in both building 1 and building 2. Notably, the response from STELLAR remains relatively stable, highlighting its ability to maintain accurate localization despite changes in the environment.

ANVIL and VITAL, although designed to address heterogeneity, show some resilience in the temporal space. However, they tend to exhibit higher localization errors, particularly when confronted with a large number of dropped APs (CI:9 and above). On the other hand, SANGRIA and WiDeep, also designed for heterogeneity, are significantly influenced by temporal variations. These frameworks struggle to maintain accuracy when confronted with changing environmental conditions over time. It is worth noting that the LT-KNN framework requires frequent re-training (after every third CI) and still fails to outperform STELLAR. This is because frequent re-calibrations (i.e., re-training) can potentially lead to overfitting in most ML models.

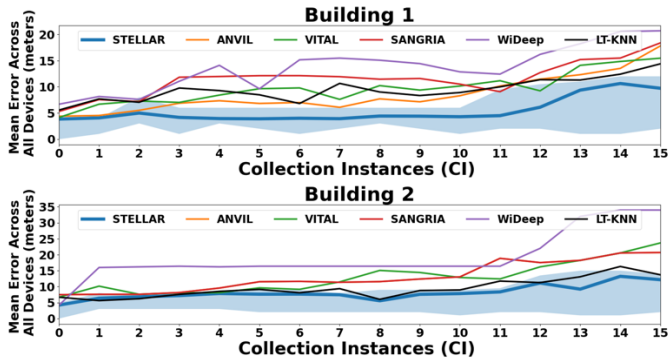

Figure 14: Average errors across all frameworks in the temporal space, blue shadow indicates best and worst-case errors for STELLAR.

To better visualize the performance disparity between STELLAR and the other frameworks, Figure 15 illustrates a series of box plots, each corresponding to a unique CI. The box plots depict the discrepancy in errors between STELLAR and its competitors for each CI. The lower whisker of the box plot denotes the smallest error difference found when comparing STELLAR with the competing frameworks, while the upper whisker indicates the highest error difference observed. The central black bar within each box represents the mean error difference. A closer examination of Figure 15 reveals an upward trend in the error difference as the CIs increase. This suggests that as temporal variations become more pronounced over time, the competing frameworks struggle to sustain uniform performance, which highlights the relative robustness of STELLAR.

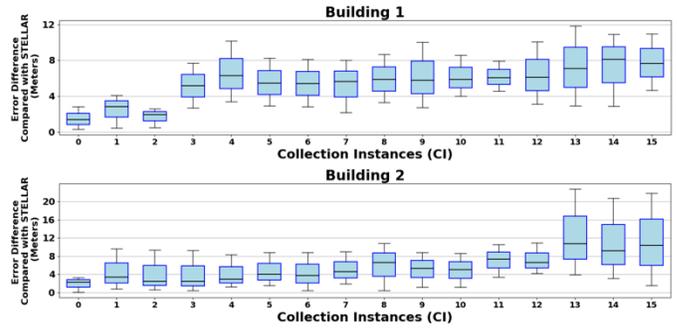

Figure 15: Box plot illustrating the minimum, maximum, and mean error differences between STELLAR and other frameworks per CI.

In conclusion, the results highlight the superior performance of the STELLAR framework in handling both heterogeneity and temporal variations. Comparing its performance with other frameworks in building 1, STELLAR demonstrates significant improvements of 75.2% over ANVIL, 80.11% over VITAL, 114.12% over SANGRIA, 153.34% over WiDeep, and 58.92% over LT-KNN. Similarly, in building 2, STELLAR outperforms ANVIL by 62.65%, VITAL by 59.89%, SANGRIA by 73.68%, WiDeep by 139.04%, and LT-KNN by 18.29%.

### C.6 Evaluation on Extended Devices and Time Period

To enhance our understanding of the STELLAR framework's resilience in handling heterogeneity and temporal variations, we conducted an extended evaluation using additional data from newer smartphone devices. In this experiment, we evaluate the performance of all the frameworks on fingerprint data collected, only by the newer smartphone devices presented in Table 2. The data was collected on the same building floorplans (building 1 and building 2) with a significant time gap of 16 months from the previously collected CI (CI:15). This temporal difference of 16 months effectively spans a period of 2 years (from CI:0), allowing us to capture and analyze data from time intervals ranging from a few seconds to 2 years. The newly collected CI, now denoted as CI:16, provides valuable insights into the evolution of the indoor environment. Notably, we observed a staggering 60% change in the APs within this time frame, representing a substantial alteration in the wireless infrastructure. To further intensify the testing and explore extended heterogeneity challenges, we incorporated newer testing devices into the evaluation. The details of these extended devices of expanded study can be found in Table 2.

Table 2: Details of extended smartphones used in our experiments.

| Manufacturer | Model | Acronym | Wi-Fi Chipset |
|---|---|---|---|
| Apple | iPhone 12 | iPhone | Qualcomm Snapdragon X55 |
| Google | Pixel 4a | Pixel | Qualcomm Snapdragon 730G |
| Nokia | 7.1 | Nokia | Qualcomm Snapdragon 836 |

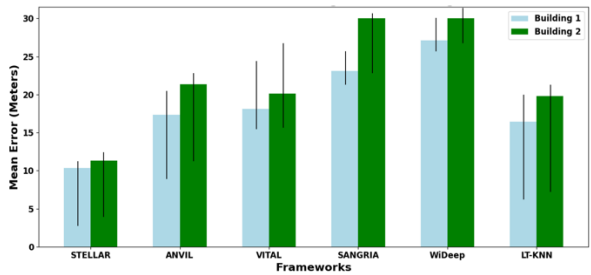

Figure 16: Average errors across all frameworks in the temporal space for extended devices with min-max intervals showing best and worst-case errors for each framework.

Figure 16 shows the performance of various frameworks on the extended devices and for CI:16. It is evident that, compared to STELLAR, all other frameworks exhibit poor performance as the intensity of temporal and heterogeneity increases, resulting in degraded performance. It is worth noting that the LT-KNN framework requires frequent re-training and, in this experiment, we re-trained LT-KNN on the previous CI (CI:15) and it still fails to outperform STELLAR. This may be because frequent re-training can potentially overfit some ML models. Additionally, it can be observed that the average error for STELLAR has also increased compared to the previous CI:15. However, it remains significantly lower than the errors produced by the state-of-the-art frameworks being compared. In building 1, STELLAR improves upon ANVIL by 68%, VITAL by 75.5%, SANGRIA by 123.93%, WiDeep by 162.69%, and LT-KNN by 59.3%. Similarly, in building 2, STELLAR achieves improves upon ANVIL by 88.6%, VITAL by 77.73%, SANGRIA by 165%, WiDeep by 165%, and LT-KNN by 74.9%.

## VII. CONCLUSIONS AND FUTURE WORKS

In this paper, we introduced STELLAR, a novel re-calibration-free framework that extends the capabilities of the ANVIL framework and demonstrates greater resilience towards heterogeneity and temporal variations in indoor localization. Through extensive experimentation, we have shown that STELLAR achieves superior localization performance across multiple heterogeneous devices and across time periods, ranging from seconds to two years. Our results indicate that STELLAR outperforms state-of-the-art localization frameworks by delivering performance improvements of 8 to 75% in terms of heterogeneity and 18 to 153% in terms of temporal variations. STELLAR also outperforms state-of-the-art localization frameworks using extended devices and timeframes by delivering performance improvements from 68 to 165%. These evaluations were conducted using real-world data and indoor environment scenarios, highlighting the effectiveness of our framework.

As part of our future work, we plan to expand our research in several key directions. Firstly, we aim to explore more complex non-rectilinear paths within indoor building environments. This will allow us to address challenges through unconventional navigation scenarios. Additionally, we recognize the importance of addressing security concerns in indoor localization systems. We plan to investigate methods to enhance the security [35] and privacy [36] aspects of indoor localization systems, ensuring that it remains robust against potential threats and vulnerabilities.

### ACKNOWLEDGEMENTS

This research is supported in part by the National Science Foundation grant CNS-2132385.

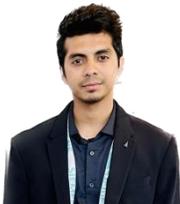

**Danish Gufran** (Student Member, IEEE) received his Engineer's degree in Electronics and Instrumentation Engineering from Visveswaraya Technological University, India, in 2020. Presently, he is pursuing his Ph.D. in Computer Engineering at Colorado State University, Fort Collins with a focus on designing deep machine learning models for embedded systems. His research is dedicated to leveraging low-power deep machine learning models to enhance indoor localization and navigation using embedded and IoT devices.

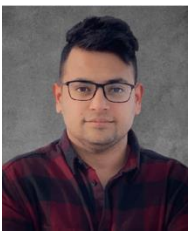

**Saideep Tiku** received his Ph.D. in Electrical Engineering from Colorado State University, Fort Collins in 2022. Presently, he is working as a Senior Systems Architect with Micron. His research interests include indoor localization and energy efficiency for fault-tolerant embedded systems. His work in the domain of machine learning-based indoor localization has been published and recognized globally in several conferences. He is the recipient of two best paper/poster awards and authored patents in the domain of machine learning-based indoor localization and other fields of embedded systems. He has co-authored one book and multiple book chapters. He is a member of the IEEE.

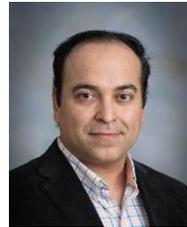

**Sudeep Pasricha** (Senior Member, IEEE), received his Ph.D. in Computer Science from the University of California, Irvine in 2008. He is currently a Walter Scott Jr. College of Engineering Professor in the Department of Electrical and Computer Engineering, at Colorado State University. His research focuses on the design of innovative software algorithms, hardware architectures, and hardware-software co-design techniques for energy-efficient, fault-tolerant, real-time, and secure computing. He has co-authored five books and published more than 300 research articles in peer-reviewed journals and conferences that have received 17 Best Paper Awards and Nominations at various IEEE and ACM conferences. He has served as General Chair and Program Committee Chair for multiple IEEE and ACM conferences and served in the Editorial board of multiple IEEE and ACM journals. He is a Senior Member of the IEEE (Computer Society), Distinguished Member of the ACM, and an ACM Distinguished Speaker.